\documentclass[10pt,twocolumn,letterpaper]{article}

\usepackage[pagenumbers]{cvpr}

\usepackage{microtype}
\usepackage{epsfig}
\usepackage{float}
\usepackage{placeins}
\usepackage{stfloats}
\usepackage{tabularx}
\usepackage{xstring}
\usepackage{multirow}
\usepackage{url}
\usepackage[hang,flushmargin]{footmisc}
\usepackage{adjustbox}
\usepackage{graphicx}

\renewcommand{\paragraph}[1]{\vspace{.5em}\noindent\textbf{#1.}}

\usepackage{xspace}

\usepackage{soul}
\setuldepth{foobar}

\newcommand{\reffig}[1]{Fig.~\ref{#1}}
\newcommand{\reftable}[1]{Tab.~\ref{#1}}
\newcommand{\refsec}[1]{Sec.~\ref{#1}}
\newcommand{\refeq}[1]{Eq.~\ref{#1}}

\definecolor{cvprblue}{rgb}{0.21,0.49,0.74}
\usepackage[pagebackref,breaklinks,colorlinks,allcolors=cvprblue]{hyperref}

\def\paperID{5029}
\def\confName{CVPR}
\def\confYear{2026}

\title{Natural Human Motion Recovery by Aligning High-Order Temporal Dynamics from Monocular Videos}

\author{
    Dingkun Wei$^{1,2,*}$ \quad
    Zehong Shen$^{2,*}$ \quad
    Yan Xia$^{3}$ \quad \\
    Georgios Pavlakos$^{3}$ \quad 
    Yujun Shen$^{2}$ \quad 
    Xiaowei Zhou$^{1}$ \\
    $^{1}$Zhejiang University \quad $^{2}$Ant Group \quad $^{3}$The University of Texas at Austin \\
}

\begin{document}
\maketitle

\begingroup
\renewcommand{\thefootnote}{\fnsymbol{footnote}}
\setcounter{footnote}{0}
\footnotetext[1]{Equal contribution; work done while D. Wei was an intern at Ant Group.}
\renewcommand{\thefootnote}{\phantom{*}}
\footnotetext[2]{The authors from Zhejiang University are affiliated with the State Key Lab of CAD\&CG. Corresponding author: Xiaowei Zhou.}
\endgroup

\begin{abstract}

Human motion recovered from monocular videos often appears overly smooth or dynamically inconsistent, even when joint positions are numerically accurate. We observe that this limitation stems from the absence of reliable high-order temporal cues—velocity and acceleration—which are essential for reconstructing motion that exhibits realistic momentum, timing, and high-frequency detail. We introduce HTD-Refine, a post-processing framework that augments existing Human Motion Recovery (HMR) pipelines using explicitly estimated high-order temporal dynamics. At the core of our system is PVA-Net, a temporal transformer that infers per-joint 2D positions, 3D velocities, and 3D accelerations directly from a monocular video. These predicted dynamics serve as soft yet informative constraints in a global optimization procedure that refines world-space trajectories, significantly reducing jitter, suppressing oversmoothing, and restoring physically plausible motion. Extensive experiments on challenging in-the-wild benchmarks show that HTD-Refine consistently improves state-of-the-art HMR methods, yielding more accurate global trajectories and substantially more natural motion dynamics. Our results highlight the critical role of high-order temporal modeling in advancing monocular human motion recovery. 
Project page: \url{https://zju3dv.github.io/htd-refine/}
\end{abstract}    
\section{Introduction}
\label{sec:intro}

Monocular world-grounded human motion recovery aims to reconstruct a person’s 3D trajectory in a global coordinate frame with dynamics faithful to real human movement. 
Achieving \emph{natural} motion, however, requires more than accurate joint positions: perceptual realism hinges on high-order temporal cues—\textbf{velocities} that evolve smoothly, \textbf{accelerations} that reflect plausible momentum, and energy profiles that convey intention, balance, and rhythm.
Recovering such dynamics from monocular video would unlock broad applications: perceptually convincing character animation and generation from casual videos~\cite{amass,t2m}, stable trajectories in humanoid imitation learning~\cite{liao2025beyondmimic,ze2025twist}, and general video-based understanding of human behavior like gait analysis and health care~\cite{skel,stenum2021applications}.

\begin{figure}
    \centering
    \includegraphics[width=0.95\columnwidth]{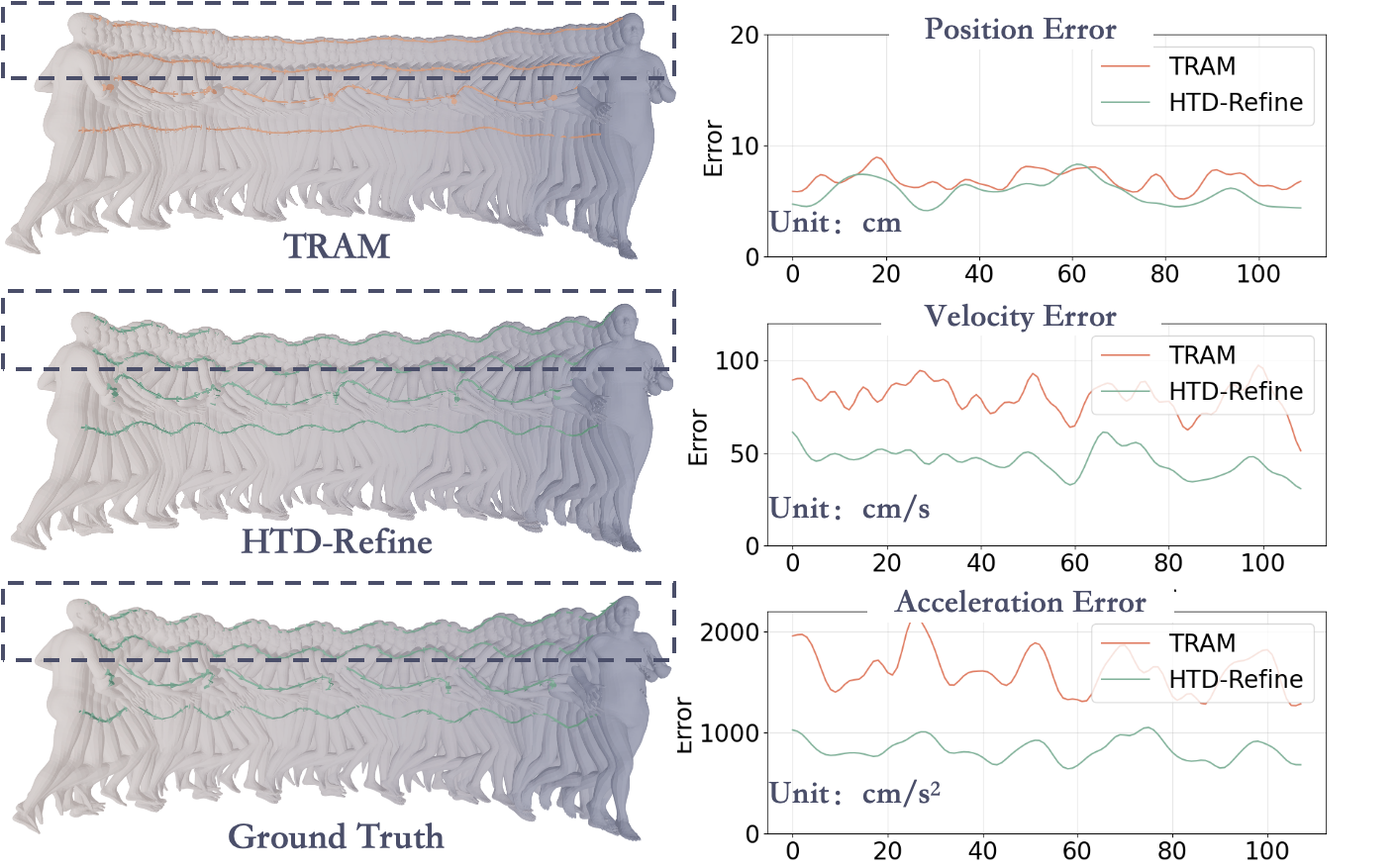}
    \caption{
        \textbf{Comparison between TRAM and TRAM + HTD-Refine.}
        TRAM achieves low position error but exhibits inconsistent high-order dynamics, while our refinement restores accurate velocities and accelerations, producing more natural motion.
    }
    \label{fig:teaser}
\end{figure}

Despite steady progress on joint-position accuracy~\cite{hmr2,nlf,tram,wham,gvhmr,prompthmr,human3r,r1_zhang2025human,r2_liu2025heuristic,r3_zhang20243d,r4_hou2024causal,r5_shi2024generating,r6_liu2024geometry}, monocular motion recovery still produces trajectories that appear either jittery~\cite{hmr2,nlf,human3r} or overly smoothed~\cite{tram,wham,gvhmr}, resulting in perceptually unnatural motion. 
This stems from the fact that human motion is extremely sensitive to small numerical errors: even minor pose deviations accumulate along the kinematic chain and noticeably degrade dynamic fidelity, regardless of low position error. 
The challenge is further exacerbated by the widespread use of low-frame-rate data (e.g., 30\,FPS) for training and evaluation, which fails to capture high-frequency transients and leads models to systematically underfit rapid motion changes.

Existing methods attempt to address these issues in two ways.  
\emph{Temporal-smoothing approaches}—including network-induced continuity in TRAM~\cite{tram}, GVHMR~\cite{gvhmr}, or autoregressive predictors like WHAM~\cite{wham}—implicitly regularize dynamics but remain too weak to recover high-frequency changes; additional filtering often suppresses legitimate motion.  
\emph{Generative priors}, such as diffusion models in RoHM~\cite{zhang2024rohm} or VAE transition priors in HuMoR~\cite{humor}, produce plausible sequences but struggle to balance global consistency with frame-level 2D evidence, leading to instability as well as computational cost.  

In this work, we introduce \textbf{HTD-Refine}, a general post-processing framework that enhances existing HMR pipelines by explicitly enforcing high-order temporal dynamics.  
Our key idea is to guide global trajectory refinement using an estimated \emph{velocity--acceleration field} directly predicted from the input video.  
To this end, we propose \textbf{PVA-Net}, a lightweight temporal model that jointly estimates 2D keypoints, per-joint velocities, and accelerations.  
PVA-Net combines ViT-based~\cite{vit} spatial features with a compact temporal transformer, decoding spatio-temporal tokens via multi-task heads to produce frame-aligned predictions.  
Given these high-order signals, HTD-Refine optimizes the initial 3D reconstruction to match the predicted velocities and accelerations while remaining consistent with 2D observations, preserving high-frequency details and correcting temporal inconsistencies, as shown in \reffig{fig:teaser}.

In summary, our contributions are threefold:
(1) We introduce \textbf{HTD-Refine}, a general post-processing framework that refines global human motion by enforcing consistency with explicitly estimated velocity and acceleration fields under 2D keypoint constraints.
(2) We propose \textbf{PVA-Net}, a lightweight temporal model that jointly predicts 2D keypoints, per-joint velocities, and accelerations from monocular videos.
(3) Through extensive experiments on challenging in-the-wild datasets, we show that our framework consistently enhances a variety of existing HMR pipelines and achieves state-of-the-art accuracy and dynamic fidelity.
Code and models will be made publicly available.
\section{Related Work}
\label{sec:related}

\paragraph{Monocular Human Pose Estimation}
Recent progress in 3D human recovery is largely built on parametric human models such as SMPL and SMPL-X~\cite{smpl,smplx}. 
Given a single image or video frame, the goal is to align a posed mesh with 2D image evidence. 
Early approaches optimize model parameters to fit 2D keypoints detected by an off-the-shelf keypoint estimator~\cite{openpose,vitpose} via reprojection error minimization~\cite{smpl}, while subsequent learning-based methods train deep networks to directly regress pose and shape from images~\cite{hmr2,cliff}, often with adversarial or iterative refinement~\cite{spin}. 
These methods greatly improve camera-frame pose accuracy, but they typically operate in image coordinates and emphasize joint-position metrics rather than high-order temporal dynamics, limiting their suitability for applications that require metrically consistent global trajectories.

\paragraph{Monocular Global Human Motion Recovery}
\label{subsec:ghmr}
Recovering human motion in a fixed world coordinate frame from monocular video has attracted increasing attention due to applications in humanoid robotics~\cite{ze2025twist,liao2025beyondmimic,he2025asap} and animation content creation~\cite{mdm,t2m,t2mgpt}. 

Existing methods can be grouped as follows.
\textbf{(1) Camera-first then global.} 
These methods estimate camera-space motion and camera extrinsics first and then transform the result to a global frame~\cite{slahmr,tram,whac}. 
SLAHMR~\cite{slahmr} jointly optimizes camera and motion to reduce depth and scale ambiguity. TRAM~\cite{tram} combines metric depth alignment~\cite{bhat2023zoedepth} and video HMR~\cite{hmr2} to improve initialization, and WHAC~\cite{whac} introduces motion priors while TokenHMR~\cite{dwivedi2024tokenhmr} uses additional prompting information to improve performance. 
These methods effectively address global scale and camera–motion entanglement, but their temporal modeling is still dominated by smoothness assumptions and low-frame-rate training data, which can miss high-frequency dynamics.
\textbf{(2) Direct global estimation.} 
To avoid camera-motion entanglement, some methods directly regress global human motion in a fixed world frame as if driving a virtual character, typically in a frame-by-frame roll-out manner~\cite{wham,gvhmr}. 
WHAM~\cite{wham} lifts 2D keypoint sequences and image features to global 3D motion and uses contact-aware refinement to mitigate foot sliding. GVHMR~\cite{gvhmr} introduces a gravity‑aligned coordinate system to improve vertical accuracy and stability.
Although these approaches produce metrically meaningful trajectories, they typically roll out motion in an autoregressive manner and rely on implicit temporal priors, which can still lead to either jittery or overly smoothed motion, especially under rapid transients.
\textbf{(3) Direct global human and scene reconstruction.}  
More recently, JOSH~\cite{JOSH2025} and Human3R~\cite{human3r} reconstruct both the human mesh and the surrounding scene from monocular video by leveraging large‑scale pretrained reconstruction models such as VGGT~\cite{wang2025vggt} to reason about context. 
By explicitly modeling context, these approaches further improve global consistency and reduce scale ambiguity, but they primarily target geometric reconstruction and do not explicitly estimate or supervise high-order temporal dynamics such as per-joint velocity and acceleration fields.

\paragraph{High-Quality Motion and Temporal Priors}
Despite the progress in monocular global motion capture, high-end industrial applications still predominantly rely on multi-camera optical motion-capture systems (e.g., OptiTrack~\cite{optitrack}) to acquire high-frame-rate, low-noise 3D trajectories with clean contacts and subtle, natural movements. 
In contrast, motion estimated from in-the-wild monocular videos, while reaching centimeter-level joint errors, often suffers from jitter, foot sliding, and implausible accelerations, highlighting a gap between positional accuracy and perceptual naturalness.

To bridge this gap, several works introduce temporal and physical priors that operate on pose sequences. 
LEMO~\cite{lemo} learns a motion smoothness prior combined with contact-aware friction terms, significantly reducing jitter and improving foot–ground stability. 
PhaseMP~\cite{shi2023phasemp} models pose transitions in the frequency domain with a phase-conditioned prior, yielding smoother spatial–temporal evolution. 
HuMoR~\cite{humor} employs a conditional VAE as a motion prior and performs test-time optimization to filter out physically implausible trajectories, while RoHM~\cite{zhang2024rohm} refines or completes rough pose sequences using diffusion-based dynamics models. 
These approaches demonstrate that priors can substantially enhance perceptual quality beyond joint-position error alone, but they typically regularize motion \emph{implicitly}: enforcing smoothness or sampling from a learned distribution can oversuppress legitimate high-frequency motion or drift from 2D image evidence. 
In contrast, our work introduces a refinement framework that explicitly estimates per-joint velocities and accelerations from monocular video and uses them to guide post-processing of existing HMR pipelines. 
By optimizing global motion to match these high-order temporal fields while remaining consistent with 2D keypoint observations, we directly target dynamic fidelity rather than relying solely on implicit smoothness or generative priors.

\begin{figure*}[t]
    \centering
    \includegraphics[width=1.0\textwidth]{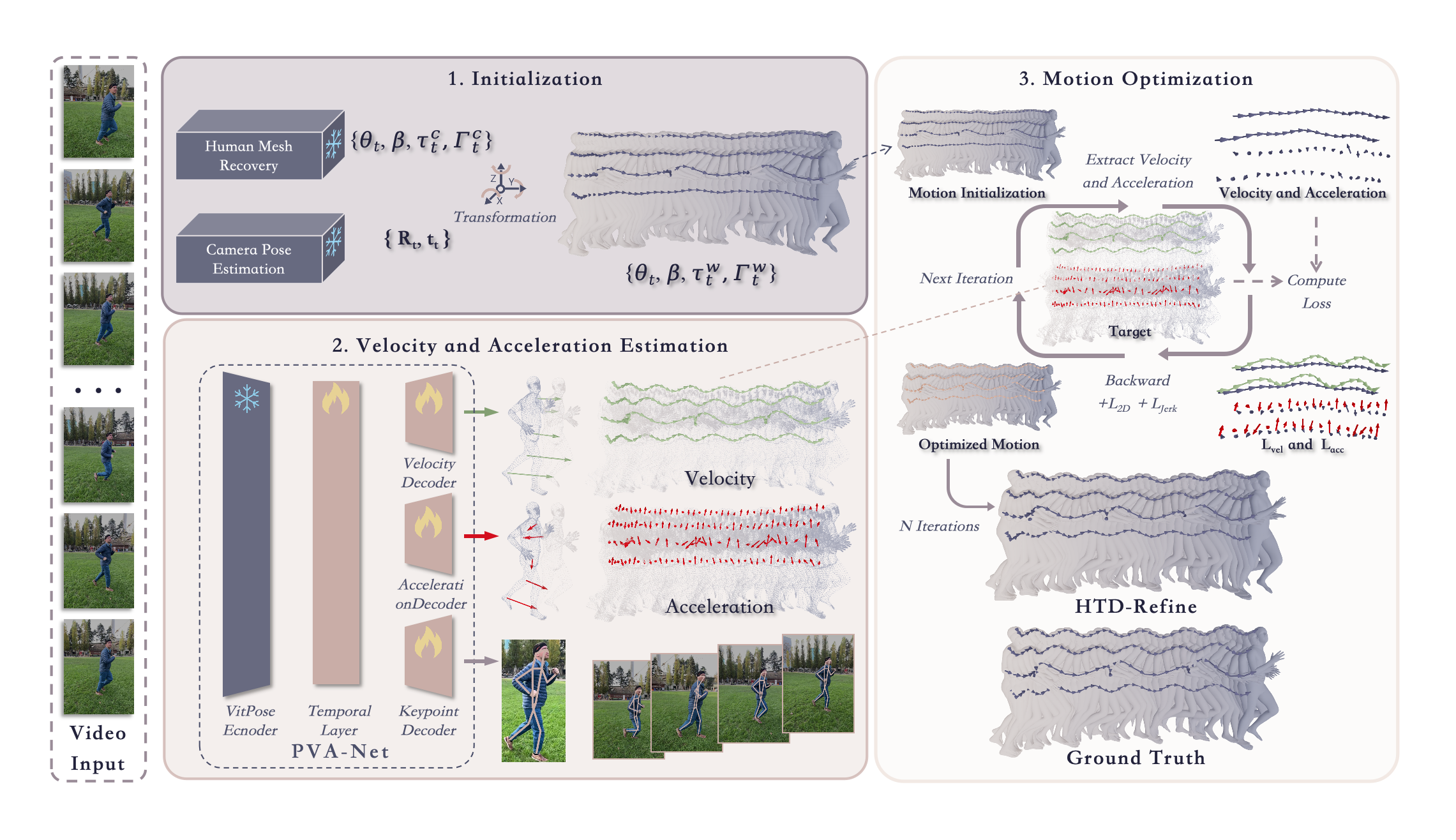}
    \vspace{-1cm}
    \caption{
        \textbf{Overview of the HTD-Refine pipeline.} Given an input video, our method proceeds in three stages. 
        (a) \textit{Initialization.} We first apply an off-the-shelf human mesh recovery model~\cite{tram,gvhmr} and a camera pose estimator~\cite{tram,droid} to obtain per-frame camera-space human pose and camera extrinsics, which are then transformed into world coordinates. 
        (b) \textit{Velocity and acceleration estimation.} In addition to predicting per-joint 2D keypoint positions, our PVA-Net also outputs camera-space 3D joint velocities and accelerations that serve as high-order motion cues.
        (c) \textit{Motion optimization.} We extract velocities and accelerations from the current global motion, impose losses against the PVA-Net predictions, and optimize the motion over $N$ iterations with additional reprojection loss and jerk loss.
    }
    \label{fig:pipeline}
\end{figure*}

\section{Method}

\label{sec:method}
Given a monocular video of length $T$, our goal is to recover natural global human motion. 
We represent global human motion using the parametric SMPL~\cite{smpl} or SMPL-X~\cite{smplx} model with per-frame parameters: 
$\{\boldsymbol{\theta}^t, \boldsymbol{\beta}, \boldsymbol{\tau}_w^{\,t}, \Gamma_w^{\,t}\}_{t=1}^{T}$,
where $\boldsymbol{\theta}^t \in \mathbb{R}^{J \times 3}$ denotes the local joint rotations, 
$\boldsymbol{\beta} \in \mathbb{R}^{10}$ is the time-invariant body shape, $\boldsymbol{\tau}_w^{\,t} \in \mathbb{R}^{3}$ is the root translation in the world frame, and $\Gamma_w^{\,t} \in \mathbb{R}^{3} $ denotes the root orientation. 
Given these parameters, the SMPL model produces mesh vertices $\mathbf{X}_w^{\,t}$ and 3D joint positions $\mathbf{J}_w^{\,t}$ in world coordinates via a differentiable forward kinematics and skinning function. 
For clarity, we use the SMPL formulation throughout, though our method applies equally to SMPL-X.

\reffig{fig:pipeline} illustrates the overall pipeline of HTD-Refine. \refsec{subsec:initialization} briefly reviews the general HMR paradigm used for initial motion estimation. Subsequently, \refsec{subsec:model} presents our PVA-Net, which predicts camera-space 3D joint velocities and accelerations. Building on this, \refsec{subsec:optimization} describes how these higher-order kinematic cues are utilized to refine the initial estimate into natural, metrically consistent global motion.

\subsection{Initialization}
\label{subsec:initialization}
Following TRAM~\cite{tram}, we express the motion as camera-space SMPL parameters $\{\boldsymbol{\theta}^t,\boldsymbol{\beta},\boldsymbol{\tau}_c^t,\Gamma_c^t\}$ 
and world-space camera extrinsics $\{\mathbf{R}_c^t,\mathbf{t}_c^t\}$.

We obtain the world-space root orientation by composing the rotations:
\begin{equation}
    \Gamma_w^t = \mathbf{R}_c^t \, \Gamma_c^t ,
\end{equation}
and transforming translations with respect to the human root offset $\mathbf{t}_{\mathrm{root}}$ in the SMPL coordinate:
\begin{equation}
    \boldsymbol{\tau}_w^t 
    = \mathbf{t}_c^t 
    + \Big(\mathbf{R}_c^t (\boldsymbol{\tau}_c^t + \mathbf{t}_{\mathrm{root}}) 
    - \mathbf{t}_{\mathrm{root}}\Big).
\end{equation}
Local joint rotations $\boldsymbol{\theta}^t$ and shape $\boldsymbol{\beta}$ remain unchanged, yielding the world-space sequence $\{\boldsymbol{\theta}^t,\boldsymbol{\beta},\boldsymbol{\tau}_w^t,\Gamma_w^t\}$.

This initialization provides a consistent world-frame trajectory but lacks high-order temporal cues, motivating our refinement stage that reinstates realistic dynamics.

\subsection{PVA-Net}
\label{subsec:model}

\paragraph{Prediction Targets}
To recover the missing high-order temporal cues, we estimate motion quantities directly in the camera coordinate system of each frame, avoiding entangling supervision with global scale ambiguity and camera egomotion.

Let $\mathbf{J}_c^t \in \mathbb{R}^{J\times 3}$ denote per-joint 3D positions expressed in the camera frame.  
We define camera-space velocities using finite differences:
\begin{equation}
    \label{formu:vel3d}
    V_c^t = \frac{\mathbf{J}_c^t - \mathbf{J}_c^{t-1}}{\Delta t}.
\end{equation}

Although velocity encodes instantaneous joint speed, it is directly proportional to the unknown global scale in monocular 3D reconstruction.
A taller subject, a closer camera, or an inaccurate focal length uniformly rescales all 3D positions; hence, the induced 3D velocity field scales proportionally.
Velocity is further contaminated by low-frequency camera drift, which appears as smooth apparent motion and is difficult to distinguish from true subject movement.

In contrast, second-order differences attenuate slowly varying global trends, such as approximately constant camera drift, while emphasizing temporally salient events including motion onset, stops, and direction reversals. Compared with absolute velocities, these high-curvature motion features are less sensitive to scale and more robust to camera motion, yielding a cleaner supervisory signal for temporal refinement.
We therefore approximate per-joint acceleration using a second-order finite difference:
\begin{equation}
    \label{formu:acc3d}
    A_c^t = \frac{\mathbf{J}_c^{t+1} - 2\mathbf{J}_c^{t} + \mathbf{J}_c^{t-1}}{(\Delta t)^2}\,.
\end{equation}

In addition, 2D keypoints remain essential cues for temporally stable HMR optimization~\cite{glamr,lemo,zhang2024rohm}.
While single-frame detectors such as ViTPose-L~\cite{vitpose} perform well on individual images, they can exhibit jitter, occlusion failure, and temporal inconsistency in videos.
We therefore introduce a temporal extension of ViTPose that produces stabilized 2D keypoints.

Building on these observations, we design a lightweight temporal transformer, PVA-Net, which jointly predicts:
\[
    \{ K^t \in \mathbb{R}^{J\times 2}\}_{t=1}^{T}  ,\;
     \{  V_c^t \in \mathbb{R}^{J\times 3}\}_{t=2}^{T} ,\;
      \{ A_c^t \in \mathbb{R}^{J\times 3} \}_{t=2}^{T-1}
\]
thereby providing consistent 2D evidence, robust high-order cues, and high-order temporal anchors to guide motion refinement.

\begin{figure}[t]
    \centering
    \includegraphics[width=0.95\columnwidth]{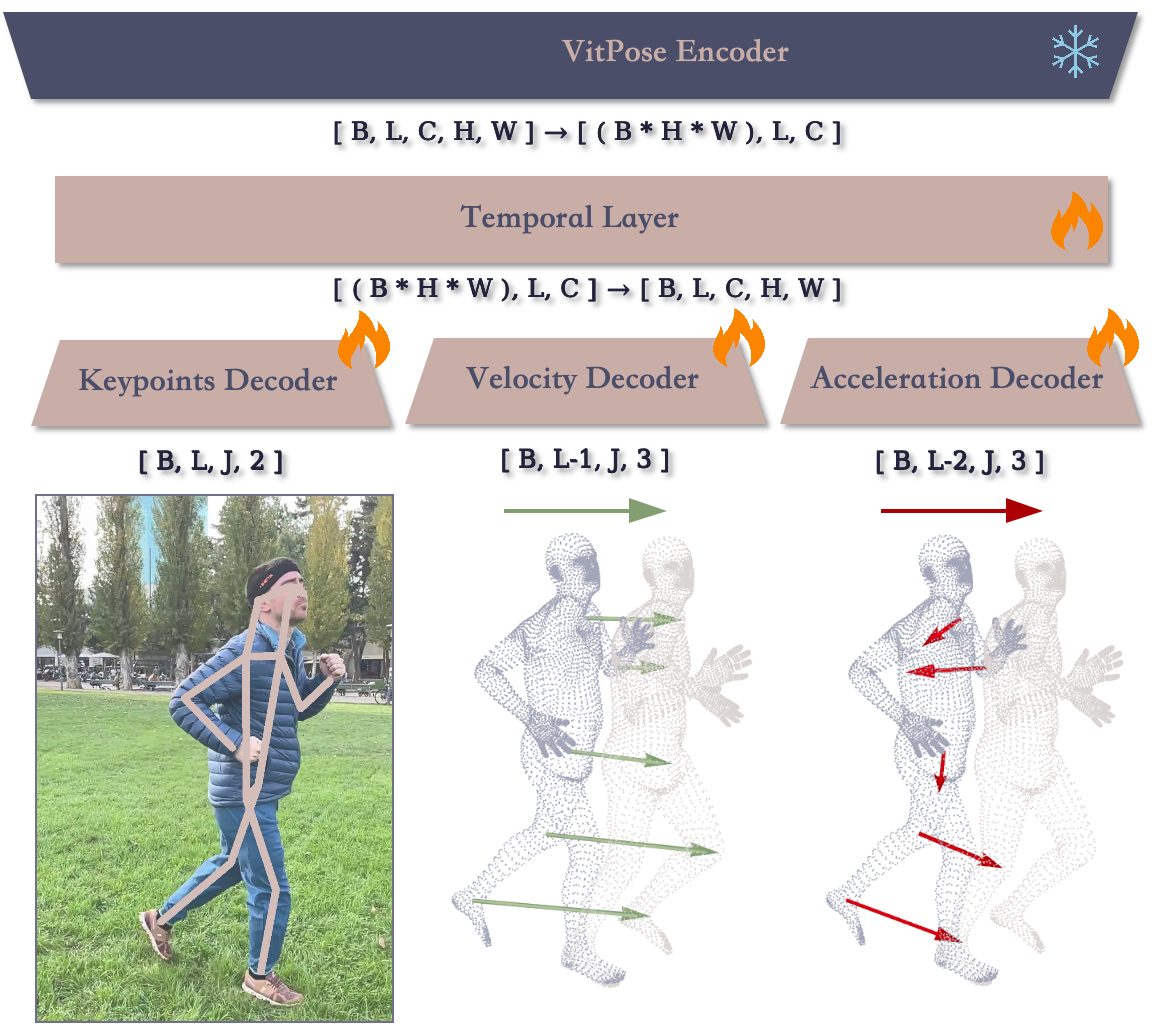}
    \caption{
        \textbf{Architecture of PVA-Net.}
        A ViTPose encoder (snowflake: frozen) extracts per-frame features, which are reshaped and processed by a lightweight temporal transformer (flame: trainable).
        Three decoders then predict per-joint keypoints, 3D velocity, and acceleration.
        The right panel visualizes the predicted velocity (blue) and acceleration (red) along the motion.
        \(B\): batch size, \(L\): frames, \(C\): channels, \(H\times W\): spatial grid, \(J\): joints.
    }
    \label{fig:pva_net}
\end{figure}

\paragraph{Network Design}
Leveraging large pose-pretrained vision encoders with modest fine-tuning is an effective strategy for transferring strong image priors to downstream motion tasks.
Following this principle, we adopt a simple and interpretable architecture, illustrated in \reffig{fig:pva_net}.
A ViT backbone initialized from ViTPose-L extracts per-frame visual features.
To model temporal evolution, we introduce a lightweight temporal transformer consisting of eight blocks.

Inspired by GVHMR~\cite{gvhmr}, which highlights the importance of accurate temporal signal modeling in downstream motion estimation, we incorporate \emph{rotary positional embeddings} (RoPE~\cite{rope}) into the temporal attention layers.
RoPE provides a continuous and geometry-aware encoding of temporal offsets, allowing the model to better capture high-order motion patterns such as motion initiation, reversals, and rhythmic cycles, while maintaining robustness to sequence length and global phase shifts.

Finally, three small prediction heads transform the spatio-temporal features into per-joint keypoints, per-joint camera-space velocity \(V_c^t\) and acceleration \(A_c^t\), producing temporally coherent high-order cues that drive our refinement module.

\paragraph{Training and Supervision}
PVA-Net is initialized from ViTPose-L~\cite{vitpose} and trained on standard video datasets with 3D human annotations (BEDLAM~\cite{bedlam}, RICH~\cite{rich}, and H36M~\cite{h36m}). 
We freeze the ViT backbone and train only the temporal transformer and the prediction heads.
The total loss is defined as:
\begin{equation}
    L_{\text{total}} = \alpha_{H} L_{\text{H}} + \alpha_{V} L_{\text{V}} + \alpha_{A} L_{\text{A}} + \alpha_{tgm} L_{\text{tgm}}
\end{equation}
with the terms specified as:
\begin{equation}
    L_{\text{V}} =  \frac{1}{T-1}  \sum_{t=2}^{T} \left\| \hat V_c^t  -  V_c^t  \right\|^2 ,
\end{equation}
\begin{equation}
    L_{\text{A}} =  \frac{1}{T-2}  \sum_{t=2}^{T-1} \left\| \hat A_c^t  -  A_c^t  \right\|^2 ,
\end{equation}
\begin{equation}
    L_{\text{H}} =  \frac{1}{T}  \sum_{t=1}^{T} \left\| \hat H^{t} -  H^{t} \right\|^2 ,
\end{equation}
\begin{equation}
    L_{\text{tgm}} =  \frac{1}{T-1}  \sum_{t=2}^{T} \left\| (\hat H^{t}- \hat  H^{t-1} ) - (  H^{t} - H^{t-1}) \right\|^2
\end{equation}
where $ V_c^t$ and $ A_c^t$ are computed using \refeq{formu:vel3d} and \refeq{formu:acc3d}, with the hat symbol ($\hat{\cdot}$) denoting the predictions from PVA-Net. $\hat H^t$ and ${H}^t$ denote the per-joint probability heatmaps. We obtain the predicted 2D keypoints $ \hat K^t$ by taking the argmax of the heatmaps. 
Inspired by VDA~\cite{vda}, we adopt a temporal gradient matching loss $L_{tgm}$ to encourage temporal consistency of the 2D keypoints.

This supervision jointly constrains the temporal module with 2D keypoints, 3D velocities and 3D accelerations, leading to temporally coherent high-order predictions.

\subsection{Optimizing Global Human Motion}
\label{subsec:optimization}
\paragraph{Input and Pre-Processing}
Our inputs comprise three parts: (1) an initialization of the global human motion in terms of SMPL parameters, where the shape $\boldsymbol{\beta}$ is kept fixed and we optimize the pose $\boldsymbol{\theta}$, global translation $\boldsymbol{\tau}$, and global orientation $\Gamma$; (2) the camera poses $\{ R_t, t_t\}_{t=1}^T$, which are used for projection; and (3) per-frame predictions from PVA-Net, including 2D keypoints $K^t$, camera-space 3D velocities $V_c^t$, and accelerations $A_c^t$.

Because depth-based scaling of camera trajectories may not align with the true scale of human motion, we apply a lightweight calibration that encourages consistency between the magnitude of the induced 3D joint velocity and the predicted magnitude, so that the initial global scale is coherent before optimization. 

\paragraph{Optimization Algorithm}
Given the initial parameters, SMPL produces world-space joints $\mathbf{J}^t$, which are projected to the image using the perspective camera model $\pi_t(\cdot)$ to obtain 2D keypoints. 
Temporal finite differences on each joint yield 3D velocities and accelerations. We then evaluate a weighted energy that balances target alignment (2D keypoints, 3D velocity, 3D acceleration) with smoothness (jerk) and parameter regularization. Gradients are back-propagated through SMPL and the projection, and we update $\{\boldsymbol{\theta}_w^t,\Gamma_w^t,\boldsymbol{\tau}_w^t\}$ using the Adam optimizer.

\paragraph{Energy Function}
The energy function we aim to minimize is defined as
\begin{equation}
\begin{split}
    E\big(\boldsymbol{\theta}_w,\boldsymbol{\tau}_w,\Gamma_w\big)
    = {} & \lambda_{V} \, E_{V} + \lambda_{A} \, E_{A}
    \\ & + \lambda_{K} \, E_{K} + \lambda_{\text{jerk}} \, E_{\text{jerk}} + \lambda_{\text{reg}} \, E_{\text{reg}} 
\end{split}
\end{equation}
with the individual terms specified as:
\begin{equation}
    E_{V} =  \frac{1}{T-1} \sum_{t=2}^{T} \left\| \mathbf{V}_c^{t} - \hat{V}_c^{t} \right\|_2^2 ,
\end{equation}
\begin{equation}
    E_{A} =  \frac{1}{T-2} \sum_{t=2}^{T-1} \left\| \mathbf{A}_c^{t} - \hat{A}_c^{t} \right\|_2^2 ,
\end{equation}
\begin{equation}
    E_{K} =  \frac{1}{T}  \sum_{t=1}^{T} \left\| \mathbf{K}^{t} - \hat{K}^{t} \right\|_2^2 ,
\end{equation}
\begin{equation}
    E_{\text{jerk}} =  \frac{1}{T-3}  \sum_{t=1}^{T-3} \left\| \mathbf{J}^{t+3} - 3\mathbf{J}^{t+2} + 3\mathbf{J}^{t+1} - \mathbf{J}^{t} \right\|_2^2 ,
\end{equation}
\begin{equation}
    E_{\text{reg}} = \frac{1}{T}  \sum_{t=1}^{T} \big(
        \left\| \boldsymbol{\theta}^{t} - \boldsymbol{\breve \theta}^{t} \right\|_2^2
        + \left\| \Gamma^{t} - \breve \Gamma^{t} \right\|_2^2
        + \left\| \boldsymbol{\tau}_{t} - \boldsymbol{\breve\tau}^{t} \right\|_2^2
    \big) .
\end{equation}
Here, $E_{V}, E_{A}, E_{K}, E_{\text{jerk}}, E_{\text{reg}}$ are, respectively, the velocity consistency, acceleration consistency, 2D keypoint constraints, jerk smoothness, and parameter regularization terms, weighted by coefficients: $\lambda_{K}=1.0, \lambda_{V}=1.0, \lambda_{A}=0.1, \lambda_{jerk}=10^4, \lambda_{reg}=10^4 $. The quantities $\mathbf{K}^t$, $\mathbf{V}_c^t$, and $\mathbf{A}_c^t$ are the 2D keypoints, 3D velocities, and 3D accelerations computed from the optimized motion, while $\hat{\mathbf{V}}_c^t$, $\hat{\mathbf{A}}_c^t$, and $\hat{\mathbf{K}}^t$ are their counterparts predicted by PVA-Net.
The regularization term $E_{\text{reg}}$ constrains the optimized SMPL parameters to remain close to their initial predictions (specified as $\breve{\;}$), preventing excessive deviation from the initial estimates.
All terms are evaluated over all joints and summed over time; for brevity, joint indices are omitted.

\paragraph{Post-Processing}
We aim to keep the feet and hands stable when they are in contact with the environment. In most cases, high-quality global motion requires precise, non-slipping contacts; even small drift at contact points can be unnatural. We use a simple velocity-based rule: given a velocity threshold $\xi_v = 0.1$, we compute a stationary probability $p_{s}$ from the predicted camera-space speed:
\begin{equation}
    p_{s} = \max \bigl(0, \; 1 - \tfrac{\| \mathbf{V}^t \|}{\xi_v} \bigr),
\end{equation}
where $\| \mathbf{V}^t \|$ denotes the velocity magnitude. We then compute target joint positions $\hat{\mathbf{J}}^t$ as
\begin{equation}
    \hat{\mathbf{J}}^t = p_{s}  \, \mathbf{J}^{t} + (1-p_{s}) \, \mathbf{J}^{t+1},
\end{equation}
where $\mathbf{J}^t$ is the human joint position at time step $t$. These targets are used in a subsequent inverse kinematics (IK) step to refine the global human motion.

\section{Experiments}
\label{sec:experiments}

\subsection{Datasets and Metrics}

\paragraph{Datasets}
Following prior work~\cite{tram,gvhmr}, we evaluate our method on two challenging in-the-wild benchmarks: RICH~\cite{rich} and EMDB~\cite{emdb}.  
To assess global performance, we report results on the RICH test set and the EMDB-2 split.  
The RICH test set contains 191 videos (59.1 minutes) captured with static cameras.  
EMDB-2 comprises 25 sequences recorded with moving cameras, totaling 24.0 minutes.

\paragraph{Metrics}
For world-coordinate evaluation, we divide each predicted global trajectory into non-overlapping 100-frame segments and align each segment to the ground truth.  
When alignment is performed over the entire segment, we report World-Aligned MPJPE (WA-MPJPE, mm).  
When alignment uses only the first two frames, we report World MPJPE (W-MPJPE, mm).  
We additionally report the Root Translation Error (RTE, m) to measure global translation accuracy, as well as motion jitter (Jitter, m/s$^{3}$) and foot sliding (FS, mm) to quantify global motion quality.

Beyond these standard metrics, we introduce two measures that directly evaluate the fidelity of reconstructed dynamics: Mean Per-Joint Velocity Error (MPJVE, m/s) and Mean Per-Joint Acceleration Error (MPJAE, m/s$^2$):
\begin{align}
    \mathrm{MPJVE} &= \frac{1}{T-1} \sum_{t=2}^{T} 
    \left\| \hat{\mathbf{V}}_c^{t} - \mathbf{V}_c^{t} \right\|_2^2, \\
    \mathrm{MPJAE} &= \frac{1}{T-2} \sum_{t=2}^{T-1}
    \left\| \hat{\mathbf{A}}_c^{t} - \mathbf{A}_c^{t} \right\|_2^2,
\end{align}
where $\mathbf{V}_c^{t}$ and $\mathbf{A}_c^{t}$ denote joint velocity and acceleration in the camera coordinate frame.

\subsection{Main Results}
We benchmark representative HMR methods~\cite{tram,gvhmr,human3r} on RICH~\cite{rich} and EMDB~\cite{emdb}, representing static and moving-camera scenarios, respectively.
For TRAM~\cite{tram} and Human3R~\cite{human3r}, we evaluate the official implementation under identical experimental conditions.
For GVHMR~\cite{gvhmr}, since camera trajectories are not available, we use the trajectories estimated by TRAM to ensure a fair comparison across methods.

\begin{table}[t]
\setlength\tabcolsep{3.6pt}
\centering
\scriptsize
\resizebox{\columnwidth}{!}{
\begin{tabular}{lccccccc}
\toprule
Model & Jitter & FS & MPJVE & MPJAE & WA-MPJPE & W-MPJPE & RTE \\
\midrule
TRAM (w/ traj filter) & 25.1 & 12.0 &0.6& 12.3& 78.8 & 221.3 & 1.5 \\
TRAM+HTD-Refine & \bf{6.6} & \bf{7.5}  &\bf{0.4}&\bf{8.0}& \bf{71.7} & \bf{204.9} & 1.5 \\
\midrule

GVHMR & 17.2 & \bf{4.0} & 0.6 & 10.4 & 118.7 &292.7 & 2.1 \\
GVHMR+HTD-Refine& \bf{7.2} & 5.7 & \bf{0.4} &\bf{7.9}& \bf{69.2} & \bf{192.4} & \bf{1.5} \\
\midrule
Human3R & 529.6 & 60.0 & 2.9 & 143.3 & 169.0  &\bf{ 367.1} &2.2 \\
Human3R+HTD-Refine & \bf{132.5} & \bf{23.2} & \bf{1.3}  & \bf{39.4} & \bf{156.2} & 391.4 & 2.2 \\

\bottomrule
\end{tabular}
}
\caption{
\textbf{Quantitative comparison on the EMDB-2~\cite{emdb} benchmark with moving cameras.}
We report global motion stability (Jitter, FS), joint dynamics (MPJVE, MPJAE), and global accuracy (WA-MPJPE, W-MPJPE, RTE) for different baselines and their variants with HTD-Refine.
HTD-Refine consistently improves motion smoothness and global accuracy over most baselines.
}
\label{tab:emdb}
\end{table}

\paragraph{Moving camera}

As shown in \reftable{tab:emdb}, HTD-Refine consistently improves motion quality across all baselines.  
Motion smoothness is markedly enhanced: jitter is significantly reduced across all baselines ($58.1\%$--$75.0\%\downarrow$). Foot sliding (FS) also decreases substantially for TRAM and Human3R ($37.5\%$--$61.3\%\downarrow$). GVHMR exhibits a marginal absolute FS increase (+1.7\,mm), which we attribute to the baseline occasionally failing to generalize to complex scenes (e.g., skateboarding) and yielding nearly stationary predictions; although this produces an artificially low FS, HTD-Refine successfully recovers accurate global trajectories with natural motion.

These smoothness gains are rooted in more faithful joint dynamics and global accuracy. MPJVE and MPJAE are consistently reduced across all baselines ($33.3\%$--$55.2\%\downarrow$ and $24.0\%$--$72.5\%\downarrow$, respectively). Similarly, WA-MPJPE decreases across the board ($7.6\%$--$41.7\%\downarrow$). W-MPJPE largely follows this downward trend ($7.4\%$--$34.3\%\downarrow$), except for a marginal increase for Human3R. Finally, RTE remains stable or improves (up to $28.6\%\downarrow$), indicating that enforcing temporal smoothness does not compromise global trajectory consistency.

\begin{figure*}[t]
    \centering
    \includegraphics[width=0.85\textwidth]{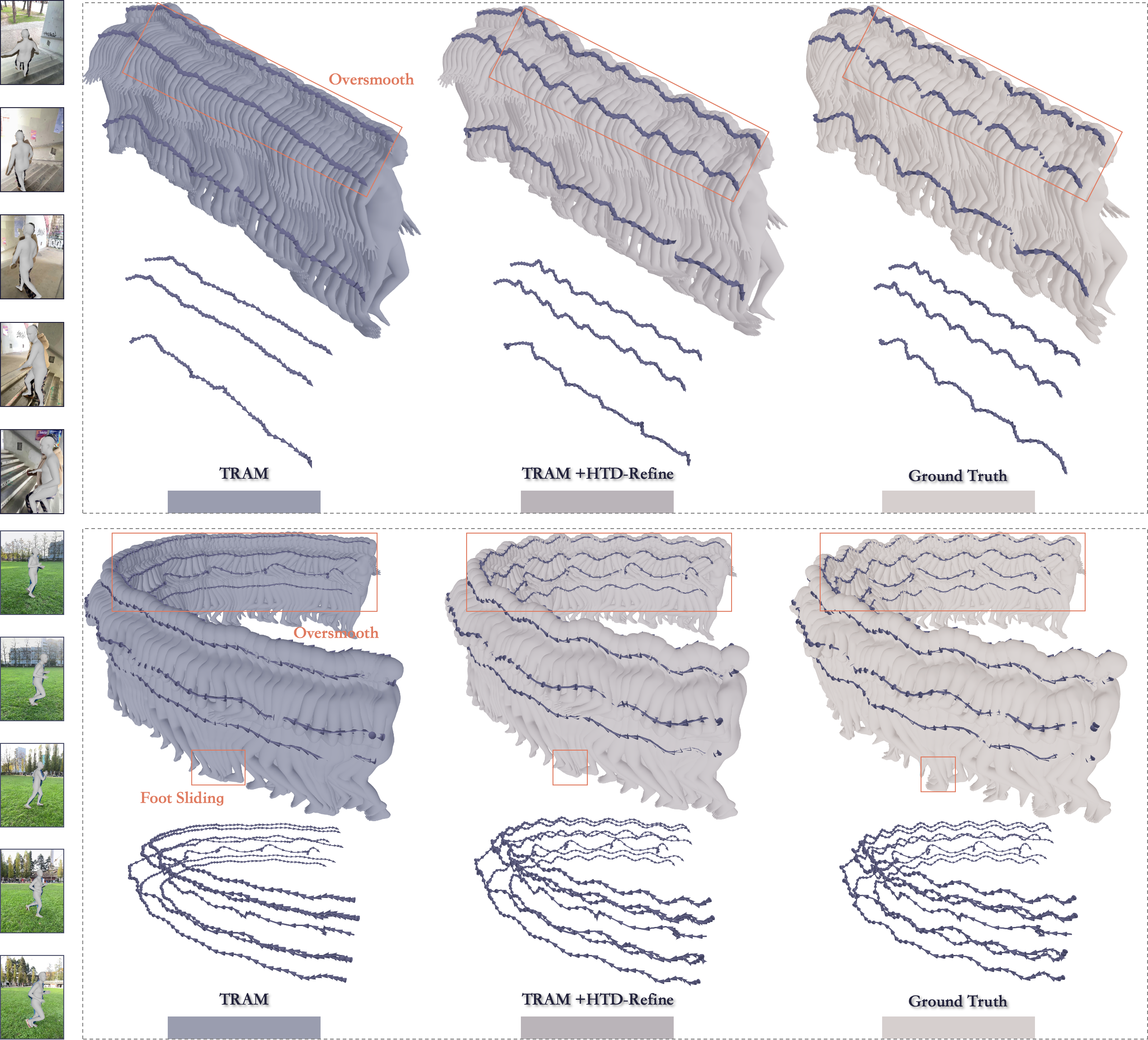}
    \caption{
        \textbf{Qualitative results on EMDB.}
        Compared to TRAM~\cite{tram}, our method substantially reduces foot sliding and over-smoothing in world space, 
        and preserves accurate camera-space poses that stay well aligned with the input video.
    }
    \label{fig:qualitive}
\end{figure*}

A qualitative comparison on EMDB~\cite{emdb} is presented in \reffig{fig:qualitive}.  
We highlight challenging cases where the baseline exhibits severe foot sliding and over-smoothed global trajectories.  
HTD-Refine markedly suppresses these artifacts, recovering world-space velocities and accelerations that better follow the ground-truth temporal dynamics, while maintaining accurate camera-space poses that remain well aligned with the input images.

\begin{table}[t]
\setlength\tabcolsep{3.6pt}
\centering
\scriptsize
\resizebox{\columnwidth}{!}{
\begin{tabular}{lccccccc}
\toprule
Model & Jitter & FS & MPJVE & MPJAE & WA-MPJPE & W-MPJPE & RTE \\
\midrule
TRAM (w/ traj filter) & 18.7 & 12.9 & 0.6 & 8.7 & 103.6 & 168.4 & 2.7 \\
TRAM+HTD-Refine & \bf{4.2} & \bf{6.5} & \bf{0.4} & \bf{5.1} & \bf{90.2} & \bf{145.3} & 2.5 \\
\midrule
GVHMR & 13.0 & 3.3 & 0.4 & 6.8 & 77.4 & 124.0 & 2.5 \\
GVHMR+HTD-Refine& \bf{3.6} & 3.3 & \bf{0.3} & \bf{4.8} & \bf{75.2} & \bf{123.8} & \bf{2.3} \\
\bottomrule
\end{tabular}
}
\caption{
    \textbf{Quantitative comparison on the RICH~\cite{rich} test set with static cameras.}
    HTD-Refine consistently enhances the temporal quality of reconstructed motion, markedly reducing jitter artifacts.
}
\label{tab:rich}
\end{table}

\paragraph{Static camera}

As shown in \reftable{tab:rich}, HTD-Refine consistently enhances motion quality across all baselines on the static-camera RICH benchmark. 
Motion smoothness improves significantly: jitter is substantially reduced across the board ($72.3\%$--$77.5\%\downarrow$). Foot sliding (FS) decreases substantially for TRAM ($49.6\%\downarrow$) and remains fully stable for GVHMR, which already exhibits very low initial sliding.

These temporal gains naturally align with more accurate joint dynamics, evidenced by consistent reductions in MPJVE ($25.0\%$--$33.3\%\downarrow$) and MPJAE ($29.4\%$--$41.4\%\downarrow$). Global accuracy also uniformly improves, reflected by across-the-board decreases in WA-MPJPE ($2.8\%$--$12.9\%\downarrow$), W-MPJPE ($0.2\%$--$13.7\%\downarrow$), and RTE ($7.4\%$--$8.0\%\downarrow$). This confirms that our refinement enhances both local smoothness and global trajectory consistency.

\begin{table*}[t]
\setlength\tabcolsep{3.6pt}
\centering
\scriptsize
\begin{tabular}{lcccccccc}
\toprule
Model & Jitter & FS & MPJVE & MPJAE & WA-MPJPE & W-MPJPE & RTE & PA-MPJPE \\ 
\midrule
TRAM & 96.0 & 18.2 & 0.8 & 27.2 & 80.6 & 223.4 & 1.5  & 36.4 \\
TRAM + Traj-Filter & 25.1 & 12.0 & 0.6 & 12.3 & 78.8 & 221.3 & 1.5 & 36.4\\
TRAM + HTD-Refine(w/o vel, w/o acc) & 10.1 & 10.5 & 0.5 & 9.8 & 73.4 & 208.0 & 1.5 & 36.7\\
TRAM + HTD-Refine(w/o acc) & 9.7 & 8.0 & 0.5 & 8.6 & 73.0 & 205.1 & 1.5 & 34.6\\
TRAM + HTD-Refine(w/o vel) & 7.9 & 8.8 & 0.5 &  8.3 & 72.5 & 205.2 & 1.5 & 35.2 \\
TRAM + HTD-Refine(w/o post-proc) & 6.5 & 7.9 & 0.4 &  8.0 & 71.2 & 204.0 & 1.5 & 34.0 \\
TRAM + HTD-Refine & 6.6 & 7.5  &0.4&8.0& 71.7 & 204.9 & 1.5 & 34.1 \\
\bottomrule
\end{tabular}
\caption{
\textbf{Ablation studies.}
We compare our method with five variants on the EMDB dataset~\cite{emdb}.
}
\label{tab:ablation}
\end{table*}

\subsection{Comparison with Refinement Methods}
\begin{table}[t]
\setlength\tabcolsep{3.6pt}
\centering
\scriptsize
\resizebox{\columnwidth}{!}{
\begin{tabular}{lccccc}
\toprule
Model & Jitter & FS & WA-MPJPE & W-MPJPE & RTE \\
\midrule
TRAM (w/o traj filter) & 96.0 & 18.2 &80.6 & 223.4 & 1.5 \\
TRAM (w/ traj filter) & 25.1 & 12.0 & 78.8 & 221.3 & 1.5 \\
TRAM+RoHM & 28.4 & 9.9  & - & - & - \\
TRAM+HTD-Refine & \bf{6.6} & \bf{7.5}  & \bf{71.7} & \bf{204.9} & 1.5 \\
\bottomrule
\end{tabular}
}
\caption{
\textbf{Comparison with refinement baselines on EMDB.}}
\label{tab:rohm}
\end{table}

We compare HTD-Refine with four motion refinement baselines: the Gaussian trajectory filter used in TRAM, RoHM~\cite{zhang2024rohm}, NeMF~\cite{nemf}, and PACE~\cite{pace}.
On EMDB, we use TRAM initialization and compare against the Gaussian filter and RoHM. To ensure fairness, since RoHM refines short clips rather than full sequences, we run HTD-Refine with the same clip-wise inference and stitching protocol.
On RICH, we use GVHMR initialization and compare against NeMF and PACE. Since PACE is not publicly available, we report the metrics from its original paper.

As shown in \reftable{tab:rohm}, HTD-Refine significantly outperforms both baseline refinement methods. 
Measured against the raw predictions, our method achieves the most substantial improvements in motion smoothness (HTD-Refine vs RoHM vs traj filter --- Jitter: $93.1\%\downarrow$ vs $70.4\%\downarrow$ vs $73.9\%\downarrow$; FS: $58.8\%\downarrow$ vs $45.6\%\downarrow$ vs $34.1\%\downarrow$). 
Notably, RoHM unexpectedly degrades jitter compared to the trajectory filter ($70.4\%\downarrow$ vs $73.9\%\downarrow$). This occurs because aligning motion to image observations under noisy camera poses introduces inconsistent per-clip corrections, amplifying inter-frame discontinuities. These results highlight that high-order temporal dynamics provide a more reliable refinement signal than filtering or clip-level priors when camera poses are noisy.

\begin{table}[t]
\setlength\tabcolsep{3.6pt}
\centering
\scriptsize
\resizebox{\columnwidth}{!}{
\begin{tabular}{lccccc}
\toprule
Model & Jitter & FS & WA-MPJPE & W-MPJPE & PA-MPJPE \\
\midrule
PACE & - & - & 197.2 & 380.0 & 49.3 \\
GVHMR+NeMF & 11.6 &17.1 & 245.5 & 414.0 &105.9\\
GVHMR+HTD-Refine & \bf{3.6} & \bf{3.3}  & \bf{75.2} & \bf{123.8} &  \bf{38.0}\\
\bottomrule
\end{tabular}
}
\caption{\textbf{Comparison with refinement baselines on RICH.}}
\label{tab:baseline_rich}
\end{table}

As shown in \reftable{tab:baseline_rich}, HTD-Refine also outperforms NeMF and PACE on RICH with GVHMR initialization. Compared with NeMF, it markedly improves motion smoothness, reducing jitter by $69.0\%$ and foot-sliding by $80.7\%$. 
It also achieves the best global trajectory and local pose accuracy, consistently outperforming both baselines across all spatial metrics.

\subsection{Ablation}
We analyze the effect of each component in HTD-Refine using five ablated variants on EMDB. Results are shown in \reftable{tab:ablation}.  
(1) \emph{TRAM vs.\ Traj-Filter.}  
Applying the Gaussian trajectory filter used in TRAM reduces jitter from 96.0$\rightarrow$25.1 but also removes high-frequency details and leaves substantial artifacts such as foot sliding, revealing the limitation of pure temporal smoothing.  
(2) \emph{w/o Vel \& Acc.}  
Relying on 2D pose is insufficient due to inherent depth ambiguity. Although 2D evidence helps the global trajectory,  optimization driven by 2D keypoints tends to distort local pose to satisfy the 2D reprojection constraints (PA-MPJPE 36.4$\rightarrow$36.7), leading to implausible poses despite low position error (e.g., foot dorsiflexion and twisted arms).
(3) \emph{w/o Acc.}  
Removing acceleration supervision affects motion continuity: jitter increases from 6.6$\rightarrow$9.7 and MPJAE from 8.0$\rightarrow$8.6, confirming that acceleration is the key signal for suppressing oscillations and correcting second-order dynamics.
(4) \emph{w/o Vel.}  
Removing velocity supervision mainly hurts high-frequency stability: jitter rises moderately (6.6$\rightarrow$7.9), and foot sliding increases most (7.5$\rightarrow$8.8), showing that velocity cues enforce first-order temporal consistency and maintain contact-phase fidelity.
(5) \emph{w/o post-proc.}  
Results reveal that post-processing reduces foot sliding ($7.9 \rightarrow 7.5$) at the cost of overall accuracy (e.g., WA-MPJPE $71.2 \rightarrow 71.7$). This is because enforcing foot locking will propagate deviations to other joints. We retain this module as an optional step because it is important for visual quality.

These results demonstrate that velocity and acceleration provide complementary constraints: velocity regulates phase-consistent motion, while acceleration stabilizes higher-order dynamics, together enabling natural and physically plausible global trajectories.

\section{Conclusion}
\label{sec:conclusion}

We presented \textbf{HTD-Refine}, a general post-processing framework that enhances monocular human motion recovery by explicitly modeling high-order temporal dynamics.  
By estimating per-joint velocities and accelerations through our proposed \textbf{PVA-Net} and enforcing these dynamics as soft constraints during global motion optimization, HTD-Refine effectively mitigates jitter, suppresses oversmoothing, and restores realistic temporal behavior in both camera-space motion and world-space trajectories.  
Extensive experiments across diverse in-the-wild benchmarks demonstrate that our method consistently improves the accuracy and perceptual quality of multiple strong HMR models, highlighting the importance of high-order cues for recovering natural human motion.

\newpage
\section{Acknowledgements}
This work was partially supported by the National Key R\&D Program of China (No. 2024YFB2809105), the Zhejiang Provincial Natural Science Foundation of China (No. LR25F020003), the Ant Group Research Intern Program, the Ant Group Postdoctoral Programme, and the Information Technology Center and State Key Lab of CAD\&CG, Zhejiang University.
{
    \small
    \bibliographystyle{ieeenat_fullname}
    \bibliography{main}
}
\clearpage
\appendix
\begin{center}
    \textbf{\Large Supplementary Material}
\end{center}
\vspace{10pt}
\section{Overview}
\label{sec:appendix_section}

In this supplementary material, we provide additional implementation details of PVA-Net (\refsec{subsec:implement}) and extended evaluations of PVA‑Net (\refsec{subsec:evalPVA}), followed by a detailed description of optimization (\refsec{subsec:implementOpt}). Additionally, we demonstrate the performance improvements of HTD-Refine over the baseline on the H36M ~\cite{h36m} dataset (\refsec{subsec:htd_h36m}). For a visual overview of our method and more qualitative results, please see the \textbf{supplementary video}.

\section{Implementation Details of PVA-Net}
\label{subsec:implement}
\begin{figure}[t]
    \centering
    \includegraphics[width=0.95\columnwidth]{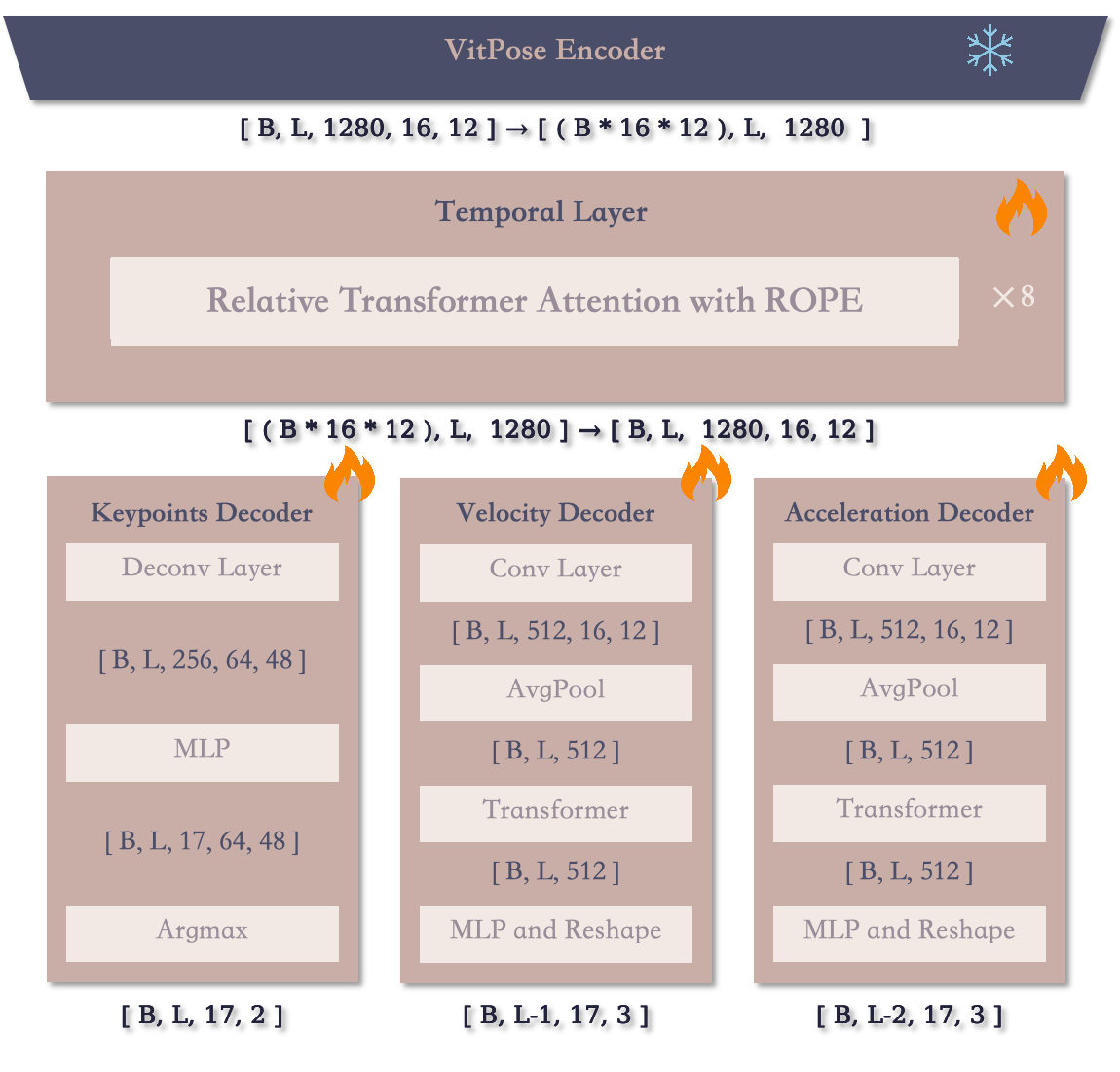}
    \caption{
        \textbf{Complete PVA-Net architecture with detailed layer specifications and input-output dimensions.}
        The system comprises: (a) ViTPose-based feature extraction, (b) 8-layer RoPE-Transformer for temporal encoding, and (c) three task-specific decoders for joint keypoints, 3D velocities, and 3D acceleration.
    }
    \label{fig:pva_full}
\end{figure}

\reffig{fig:pva_full} presents the full architecture of PVA-Net with input–output dimensions. The frame-level tokens extracted by the ViTPose encoder are fed into an 8-layer Transformer decoder that employs rotary positional embeddings (RoPE) to capture temporal dependencies across the motion sequence.
We design three decoders to handle different outputs:
Following the ViTPose design, our keypoint decoder employs a deconvolution layer for feature map upsampling, after which an MLP produces probability heatmaps for all 17 joints. The 2D joint locations are obtained via argmax;
The velocity decoder processes sequential features through a convolutional layer followed by spatial average pooling. A Transformer module then captures temporal dynamics, after which an MLP projects the features into 3D velocity vectors for all 17 joints;
The acceleration decoder adopts an identical architecture to the velocity decoder, maintaining the same processing pipeline while outputting 3D acceleration for the 17 joints.

We train PVA-Net on a mixed dataset of Human3.6M~\cite{h36m}, BEDLAM~\cite{bedlam}, and RICH~\cite{rich}, normalizing targets with dataset-specific mean and standard deviation. 
The model is optimized with AdamW (initial learning rate 2e-4) under mixed-precision (FP16). We use a global batch size of 48 and 120-frame clips. A stepwise schedule halves the learning rate at epochs 40, 60, and 80. Training on 8×NVIDIA H20 GPUs for 100 epochs leads to convergence in approximately 48 hours. With a single A6000 GPU, PVA-Net inference operates at 30+ FPS, and requires approximately 300+ GFLOPs per frame.

\section{Evaluation of PVA-Net}
\label{subsec:evalPVA}
\begin{figure*}[t]
    \centering
    \includegraphics[width=1.0\textwidth]{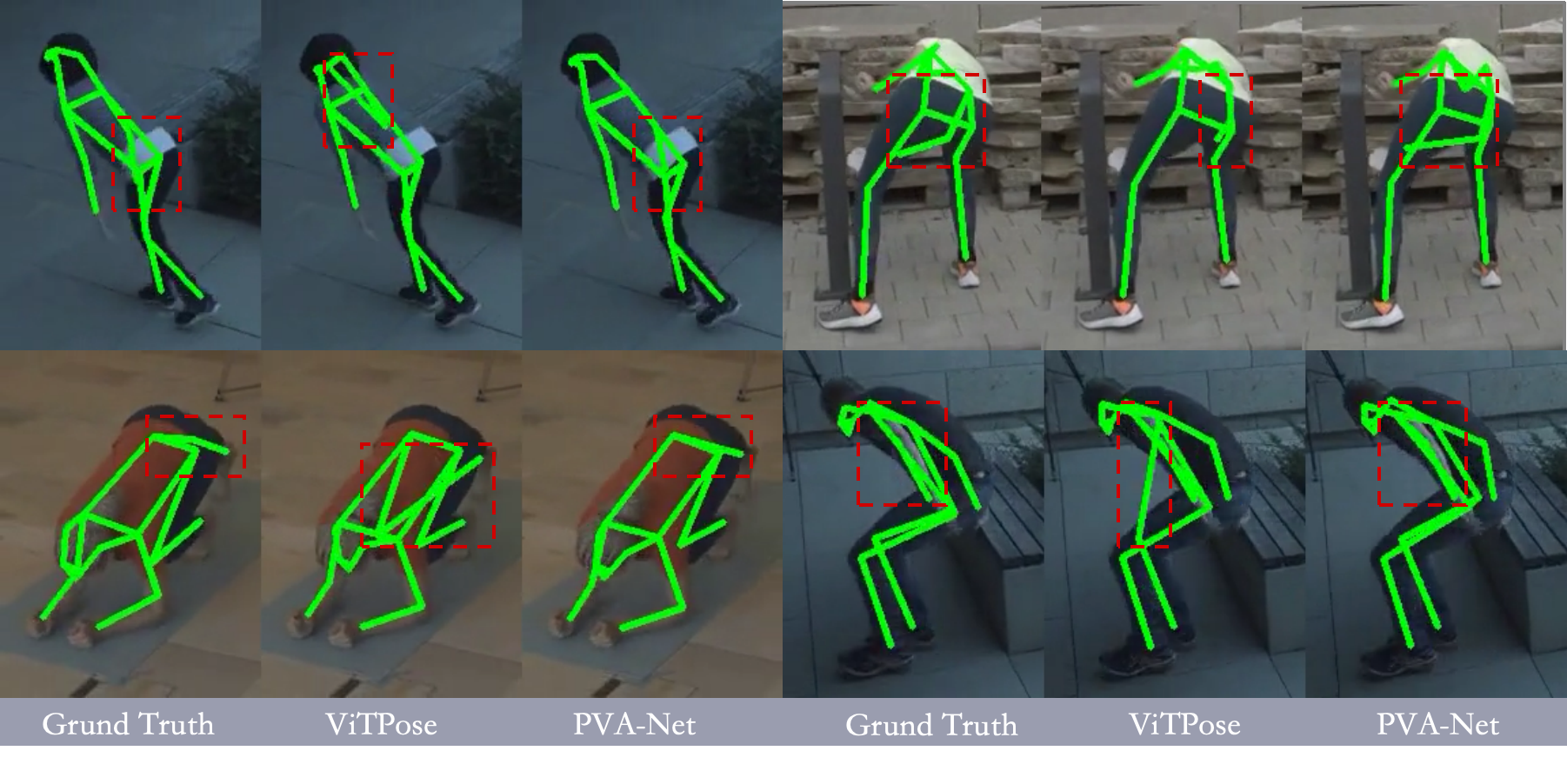}
    \caption{
        \textbf{Qualitative results of PVA-Net.} 
        Our method demonstrates enhanced robustness to occlusions by effectively leveraging temporal constraints from adjacent frames to infer plausible joint positions.
    }
    \label{fig:qualitive_pva}
\end{figure*}
\paragraph{Velocity and Acceleration}
We evaluate the performance of PVA-Net on both the RICH and EMDB datasets.
To comprehensively assess the accuracy of the predicted velocities and accelerations, we use the Percentage of Correct Estimates (PCE) metric under three progressively stricter thresholds.

Specifically, we define a dynamic value range $[r_{\min}, r_{\max}]$ based on the empirical 0.1 and 99.9 percentiles of the ground-truth distribution, computed from the BEDLAM~\cite{bedlam}, RICH~\cite{rich} and Human3.6M~\cite{h36m} datasets:
\begin{equation}
    r_{\min} = \mathrm{Percentile}_{0.1}(\mathcal{{D}}), 
    \qquad
    r_{\max} = \mathrm{Percentile}_{99.9}(\mathcal{{D}}),
\end{equation}
where $\mathcal{D}$ denotes the ground-truth velocity/acceleration statistics.
A prediction is considered correct if its absolute error is below $10\%$, $5\%$, or $1\%$ of this dynamic range, corresponding to PCE@0.10, PCE@0.05, and PCE@0.01, respectively.
For completeness, the PCE metric at threshold $\tau$ is defined as:
\begin{equation}
    \mathrm{PCE}@\tau 
    = 
    \frac{1}{N}
    \sum_{i=1}^{N}
    \mathbb{1}\!\left(
        | \mathcal{\hat{D}}_i - \mathcal{{D}}_i | 
        <
        \tau \, ( r_{\max} - r_{\min} )
    \right),
\end{equation}
where $\mathcal{{D}}_i $ and $\mathcal{\hat{D}}_i$ denote the ground-truth and predicted values, and $N$ is the number of evaluations.

\begin{table}[t]
\setlength\tabcolsep{3.6pt}
\centering
\scriptsize
\resizebox{\columnwidth}{!}{
\begin{tabular}{ccccc}
\toprule
\textbf{Dataset} & \textbf{Target} & \textbf{PCE@0.10} & \textbf{PCE@0.05} & \textbf{PCE@0.01} \\
\midrule
\multirow{2}{*}{EMDB} & Velocity & 98.2 & 93.0 & 68.2 \\
 & Acceleration & 99.6 & 98.4 & 82.3 \\
\addlinespace
\midrule
\multirow{2}{*}{RICH} & Velocity & 99.9 & 98.7 & 81.9 \\
 & Acceleration & 100.0 & 99.7 & 89.1 \\

\bottomrule
\end{tabular}
}
\caption{
    \textbf{Velocity and Acceleration Prediction Performance.}
PVA-Net achieves excellent accuracy on both EMDB and RICH datasets across PCE@0.10, PCE@0.05, and PCE@0.01  metrics, demonstrating robust performance for both velocity and acceleration estimation.
}
\label{tab:PCE_comparison}
\end{table}

As shown in \reftable{tab:PCE_comparison}, PVA-Net demonstrates strong performance across both datasets and evaluation metrics. Several key observations emerge: First, acceleration predictions consistently outperform velocity predictions, achieving higher accuracy scores, which aligns with our emphasis on acceleration modeling in the main paper. Second, we observe a performance gap between RICH and EMDB datasets (better on RICH). This discrepancy primarily stems from the inherent coupling of camera and human motion in EMDB, which presents additional challenges for estimation. These results validate that PVA-Net effectively learns discriminative velocity and acceleration representations for 3D human pose dynamics.

\paragraph{Keypoint}
In our keypoint detection comparison, to ensure a fair evaluation, we fine-tune ViTPose-L ~\cite{vitpose} on BEDLAM, RICH, and Human3.6M datasets. We employ two primary metrics for performance assessment: 
PCK@10 and PCK@05, which measure the percentage of correct keypoints within 10 pixels and 5 pixels of ground truth positions, respectively. 
Additionally, to demonstrate the temporal stability of our method, we evaluate the acceleration error (ACCEL) between predicted 2D keypoints and ground truth annotations.

\begin{table}[t]
\setlength\tabcolsep{3.6pt}
\centering
\scriptsize
\resizebox{\columnwidth}{!}{
\begin{tabular}{ccccc}
\toprule
\textbf{Dataset} & \textbf{Method} & \textbf{PCK@10} & \textbf{PCK@05} & \textbf{ACCEL}  \\
\midrule
\multirow{3}{*}{EMDB} & ViTPose-L (w/o refine) & 95.5 & 78.8 & 4.9 \\
 & ViTPose-L (w/ refine)  & 97.0 & 85.0 & 4.5  \\
 & PVA-Net & 98.3 & 88.8 & 3.7  \\
\addlinespace
\midrule
\multirow{3}{*}{RICH} & ViTPose-L (w/o refine) & 91.1 & 71.5 & 2.0 \\
 & ViTPose-L (w/ refine)  & 94.5 & 83.1 & 2.0 \\
 & PVA-Net & 97.0 & 87.7 & 0.9 \\
\bottomrule
\end{tabular}
}
\caption{
    \textbf{2D Keypoint Detection Performance on EMDB and RICH.}
    Evaluation on EMDB and RICH datasets shows that our PVA-Net achieves the best performance across all metrics (PCK@10, PCK@05, ACCEL).
}
\label{tab:2d_comparison}
\end{table}

The experimental results presented in \reftable{tab:2d_comparison} demonstrate the superior performance of our method across multiple datasets. Our approach achieves higher PCK accuracy while reducing acceleration errors, indicating improvements in both spatial precision and temporal consistency. This performance gain is primarily due to the temporal layer architecture that incorporates temporal constraints, resulting in more stable predictions. 

Additionally, \reffig{fig:qualitive_pva} further demonstrates that our method effectively resolves occlusion-induced ambiguities through inter-frame temporal reasoning.
In the \textbf{supplementary video}, we provide additional qualitative results showing that PVA-Net achieves substantially improved stability and accuracy compared to ViTPose-L across diverse motion scenarios.

\section{Implementation Details of Optimization}
\label{subsec:implementOpt}
We optimize the global SMPL parameters end-to-end using the Adam optimizer with an initial learning rate of 1e{-}3 for 1500 epochs. The learning rate is linearly warmed up during the first 10 epochs to stabilize training, and reduced by a factor of 10 at epoch 1000. 
With our fully parallelized implementation, the optimization for a single video completes in approximately 2 minutes on an NVIDIA A6000 GPU, and the runtime remains nearly constant regardless of video length, since all frames are optimized jointly in a batched formulation.

\section{Evaluation of HTD-Refine on H36M}
\label{subsec:htd_h36m}
We benchmark HTD-Refine using representative HMR methods, TRAM~\cite{tram} and GVHMR~\cite{gvhmr}, on the H36M dataset~\cite{h36m}.
\begin{table}[!ht]
\setlength\tabcolsep{1pt}
\centering
\scriptsize
% \vspace{-0.25cm}
% \resizebox{0.8\linewidth}{!}{%
\begin{tabular}{lccccccc}
\toprule
\textbf{H36M Results} & Jitter & FS & MPJVE & MPJAE & WA-MPJPE & W-MPJPE & RTE \\
\midrule
TRAM (w/ traj filter) & 10.4 & 8.4 & 0.3 & 4.9 & 77.7 & 141.7 & 1.5 \\
TRAM+HTD-Refine & \bf{4.0} & \bf{2.9} & \bf{0.2} & \bf{3.1} & \bf{69.8} & \bf{121.4} & \bf{1.4} \\
\midrule
GVHMR & 10.9 & 2.1 & 0.2 & 4.7 & 51.1 & 84.3 & 1.4 \\
GVHMR+HTD-Refine& \bf{3.6} & \bf{2.0 }& 0.2 & \bf{2.7} & \bf{44.4} & \bf{68.6} & \bf{0.8} \\

\bottomrule
\end{tabular}
% }
% \vspace{-0.65cm}
\caption{
    \textbf{Quantitative comparison on the H36M test set.}
    }
\label{tab:h36m}
\end{table}

As shown, HTD-Refine consistently improves performance for both TRAM and GVHMR across all metrics. Specifically, it significantly reduces temporal jitter and global positioning errors, while also suppressing artifacts like foot sliding. These results are consistent with our experiments on EMDB and RICH, further validating the effectiveness and plug-and-play capability of our module.

\section{Limitations and Future Work}
While HTD-Refine demonstrates effective enhancement over existing methods, we acknowledge several challenges for future work. First, our approach relies on the outputs of camera and human pose estimation, which can introduce errors into the optimization process. 
Second, the generalization ability of PVA-Net is limited in extreme scenarios underrepresented in the training data, such as skateboarding. 
In addition, under severe occlusion or motion blur, decreased confidence of PVA-Net causes the optimization to rely on regularization priors, tending to retain the initialization.
Finally, although our method jointly optimizes full video sequences, the computational overhead is still significant enough to limit real-time use. Future work may explore more efficient or streaming-based solutions.

\end{document}